\documentclass{article}
\usepackage{ijcai19}

\usepackage{times}
\usepackage{soul}
\usepackage{url}
\usepackage[hidelinks]{hyperref}
\usepackage[utf8]{inputenc}
\usepackage[small]{caption}
\usepackage{graphicx}
\usepackage{amsmath}
\usepackage{booktabs}
\usepackage{algorithm}
\usepackage{algorithmic}
\usepackage{graphicx}
\usepackage{url}
\usepackage{multirow}

\title{Generating Multiple Diverse Responses with Multi-Mapping and Posterior Mapping Selection}

\author{
Chaotao Chen$^1$\and
Jinhua Peng$^1$\and
Fan Wang$^1$\and
Jun Xu$^2$\And
Hua Wu$^1$\\
\affiliations
$^1$Baidu Inc., China\\
$^2$Harbin Institute of Technology, China\\
\emails
\{chenchaotao, pengjinhua, wangfan04, wu\_hua\}@baidu.com,
jxu@ir.hit.edu.cn
}

\begin{document}

\maketitle

\begin{abstract}

In human conversation an input post is open to multiple potential responses, which is typically regarded as a one-to-many problem. Promising approaches mainly incorporate multiple latent mechanisms to build the one-to-many relationship. However, without accurate selection of the latent mechanism corresponding to the target response during training, these methods suffer from a rough optimization of latent mechanisms. In this paper, we propose a \textit{multi-mapping} mechanism to better capture the one-to-many relationship, where multiple mapping modules are employed as latent mechanisms to model the semantic mappings from an input post to its diverse responses. For accurate optimization of latent mechanisms, a \textit{posterior mapping selection} module is designed to select the corresponding mapping module according to the target response for further optimization. We also introduce an auxiliary \textit{matching loss} to facilitate the optimization of posterior mapping selection. Empirical results demonstrate the superiority of our model in generating multiple diverse and informative responses over the state-of-the-art methods.

\end{abstract}

\section{Introduction}

Recently, generative models built upon sequence-to-sequence (Seq2Seq) framework \cite{SutskeverVL14,ShangLL15} have achieved encouraging performance in open-domain conversation, with their simplicity in learning the mapping from input post to its response directly. 
However, an input post in human conversation is open to multiple potential responses, which is typically regarded as a one-to-many problem.
Modeling diverse responding regularities as a one-to-one mapping, the Seq2Seq models inevitably favor general and trivial responses \cite{LiGBGD16}. 
Thus the rich and diverse content in human conversation can not be captured.

\begin{figure}
    \centering
    \includegraphics[scale=0.45]{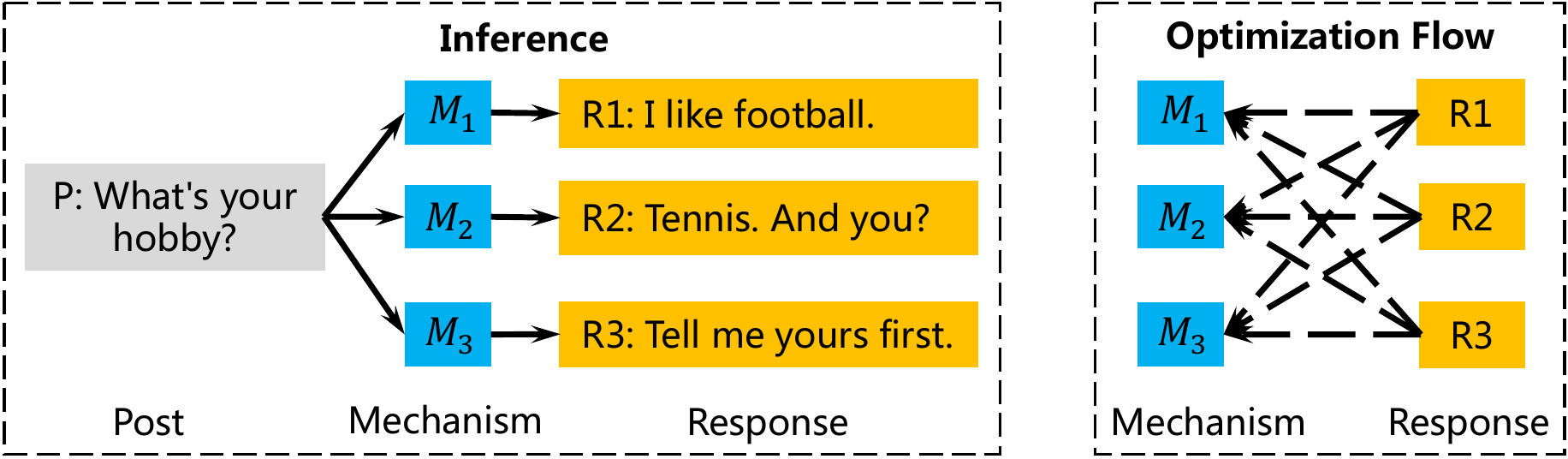}
    \caption{Overview of models with multiple latent mechanisms.}
    \label{fig:problem}
\end{figure}

To address this problem, work from \cite{ZhaoZE17,SerbanSLCPCB17} combines Seq2Seq with Conditional Variational Auto-Encoder (CVAE) and introduces a Gaussian latent distribution to build the one-to-many relationship. 
By drawing samples from the Gaussian latent distribution, multiple responses can be generated. 
However, the Gaussian latent distribution is not compatible to the multi-modal\footnote{Multi-modal means the property with multiple modes.} nature of diverse responses and lack of interpretability.

For these issues, recent approaches \cite{ZhouLCLCH17,ZhouLXLCH18,TaoGSWZY18,gu2018dialogwae} resort to incorporation of multiple latent mechanisms, each of which could model various responding regularities for an input post, as shown in Figure \ref{fig:problem}. 
For example, Zhou \textit{et al.}\ \shortcite{ZhouLCLCH17,ZhouLXLCH18} introduce multiple latent embeddings as language responding mechanisms into Seq2Seq framework, and output responses of different language styles by choosing different embeddings; Tao \textit{et al.} \shortcite{TaoGSWZY18} augment Seq2Seq model with multi-head attention mechanism, and generate responses that focus on specific semantic parts of the input post with different heads of attention.
Although these methods have shown potential to capture the multi-modal nature of diverse responses, they still fail to fulfill the one-to-many relationship, due to their inaccurate optimization of latent mechanisms.
As shown in Figure \ref{fig:problem}, given a target response, the optimization is distributed to each latent mechanism.
However, for more accurate modeling, we assume only the latent mechanism corresponding to the target response should be selected for optimization.
For example, given a questioning response, we should only optimize the latent mechanism that models interrogative responding regularities rather than the other irrelevant ones. 
Although in some methods the optimization to each latent mechanism is guided by a weight from the input post, the weight is inaccurate to represent the selection of the corresponding latent mechanism, considering the semantic gap between the input post and the target response.
With such a rough optimization, the latent mechanisms are not guaranteed to capture the diverse responding regularities.

In this paper, in order to capture the one-to-many relationship, we propose to augment the Seq2Seq framework with a \textit{multi-mapping} mechanism, employing multiple mapping modules as latent mechanisms to model the distinct semantic mappings between an input post and its diverse responses.
More importantly, to avoid the rough optimization of latent mechanisms in previous methods, in training time we incorporate a \textit{posterior mapping selection} module to select the corresponding mapping module according to the target response.
By explicitly leveraging the information in the target response (i.e.\ posterior information), it is easier to select the accurate mapping module. 
Then only the selected mapping module is updated given the target response. 
Moreover, to facilitate the optimization of posterior mapping selection, we further propose an auxiliary \textit{matching loss} that evaluates the relevance of post-response pair. 
Compared with the simple embedding mechanism and the multi-head attention mechanism whose diversity is limited to the semantic attentions in the input post, the proposed multi-mapping mechanism is more flexible to model different responding regularities. 
And it also introduces multi-modal capacity and interpretability over the Gaussian latent distribution. 
With the posterior mapping selection to ensure the accurate optimization of mapping modules, our model is more effective to capture diverse and reasonable responding regularities.

Our contributions can be summarized as follow:
\begin{itemize}
    \item We propose a multi-mapping mechanism to capture the one-to-many relationship with multiple mapping modules as latent mechanisms, which is more flexible and interpretable over previous methods.
    \item We propose a novel posterior mapping selection module to select the corresponding mapping module according to the target response during training, so that more accurate optimization of latent mechanisms is ensured. An auxiliary matching loss is also introduced to facilitate the optimization of posterior mapping selection.
    \item We empirically demonstrate that the proposed multi-mapping mechanism indeed captures distinct responding regularities in conversation. We also show that the proposed model can generate multiple diverse, fluent and informative responses, which obviously surpasses the other existing methods.
\end{itemize}

\section{Model}

\subsection{Model Overview}

Following the conventional setting for generative conversation models \cite{ShangLL15,LiMJ16}, we focus on the single-round open-domain conversation.
Formally, given an  input post $X=(x_{1},x_{2},...,x_{T})$, the model should generate a natural and meaningful response $Y=(y_{1},y_{2},...,y_{T'})$.

\begin{figure}
    \centering
    \includegraphics[scale=0.332]{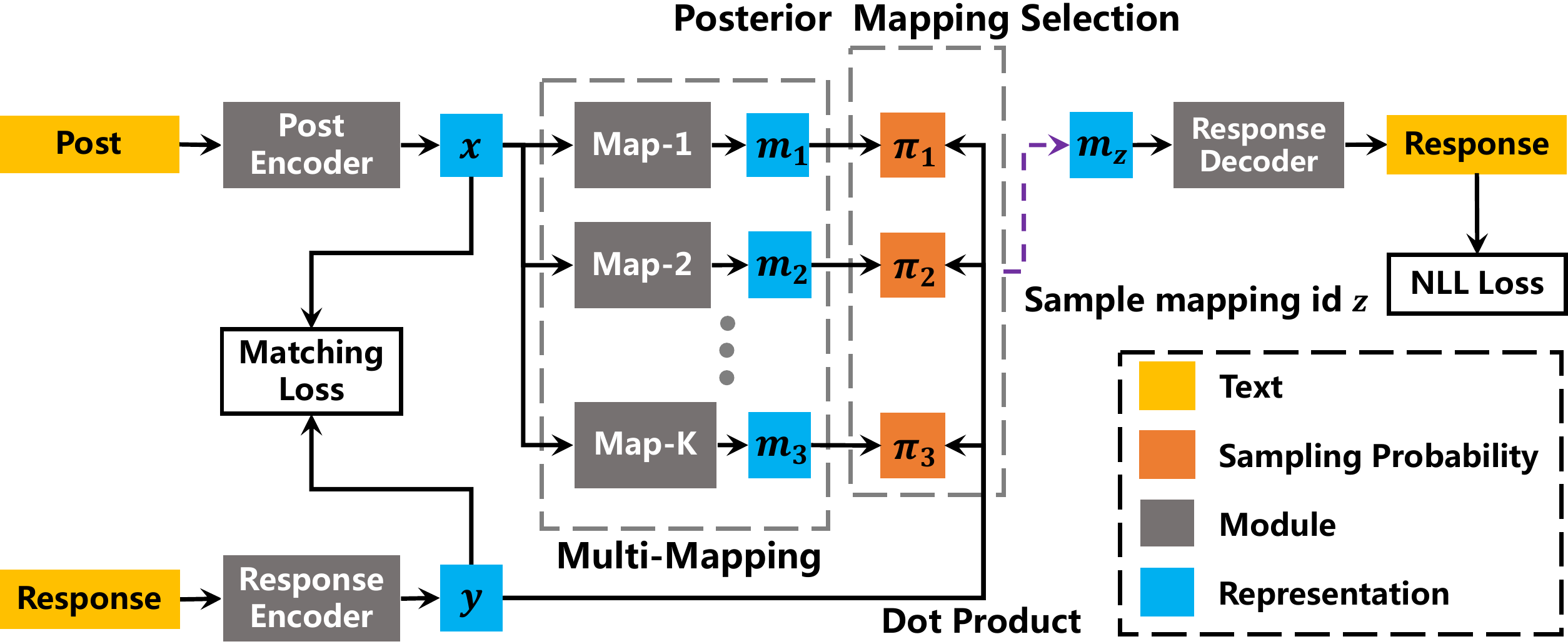}
    \caption{An overview of the proposed Seq2Seq model with multi-mapping and posterior mapping selection.}
    \label{fig:model-overview}
\end{figure}

To address the one-to-many problem in conversation, we propose a novel generative model with multi-mapping mechanism and posterior mapping selection module. 
The multi-mapping mechanism employs multiple mapping modules to capture the various underlying responding regularities between an input post and its diverse responses. 
The posterior mapping selection module leverages the posterior information in target response to identify which mapping module should be updated, so as to avoid the rough optimization in previous methods. 
The architecture of our model is illustrated in Figure \ref{fig:model-overview} and it consists of following major components:

\noindent \textbf{Post Encoder} encodes the input post $X$ into a semantic representation $\boldsymbol{x}$ and feeds it into different mapping modules.

\noindent \textbf{Response Encoder} encodes the target response $Y$ into a semantic representation $\boldsymbol{y}$ for posterior mapping selection.

\noindent \textbf{Multi-Mapping} consists of $K$ mapping modules and maps post representation $\boldsymbol{x}$ to different candidate response representation $\{\boldsymbol{m}_{k}\}^{K}_{k=1}$ through different mapping module $M_{k}$, respectively.

\noindent \textbf{Posterior Mapping Selection} selects the $z$-th mapping module that is corresponding to the target response in the training time.

\noindent \textbf{Response Decoder} generates the response based on the candidate response representation $\boldsymbol{m}_{k}$.

\subsection{Encoder}

The post encoder employs a one-layer bidirectional Gated Recurrent Unit (GRU) \cite{ChoMGBBSB14} to transform the input post $X$ into a sequence of hidden state $\boldsymbol{h}_{t}$ as follows:
\begin{align}
    \boldsymbol{h}_{t} & =[\overrightarrow{\boldsymbol{h}}_{t};\overleftarrow{\boldsymbol{h}}_{t}]\\
    \overrightarrow{\boldsymbol{h}}_{t} & =\mathbf{GRU}(\overrightarrow{\boldsymbol{h}}_{t-1},\boldsymbol{e}(x_{t}))\\
    \overleftarrow{\boldsymbol{h}}_{t} & =\mathbf{GRU}(\overleftarrow{\boldsymbol{h}}_{t+1},\boldsymbol{e}(x_{t}))
\end{align}
where $[\cdot;\cdot]$ denotes the concatenation of states, $\overrightarrow{\boldsymbol{h}}_{t}$ and $\overleftarrow{\boldsymbol{h}}_{t}$ are the forward and backward hidden states at time $t$, $\boldsymbol{e}(x_{t})$ is the embedding of word $x_{t}$. The semantic representation of input post is summarized as $\boldsymbol{x}=[\overrightarrow{\boldsymbol{h}}_{T};\overleftarrow{\boldsymbol{h}}_{1}]$.

The response encoder, which encodes the target response $Y$ to the semantic representation $\boldsymbol{y}$, follows the same structure as post encoder but with different learnable parameters.

\subsection{Multi-Mapping}
For one-to-many relationship, we introduce a multi-mapping mechanism to capture the different responding regularities, with multiple mapping modules bridging the post encoder and response decoder.
Specifically, we employ a linear mapping function as the mapping module for simplicity and leave more advanced mapping structures as future work.
Formally, the model maps the post representation $\boldsymbol{x}$
to the $K$ different candidate response representations $\{\boldsymbol{m}_{k}\}^{K}_{k=1}$ through different mapping modules as follows:
\begin{equation}
    \boldsymbol{m}_{k}=\boldsymbol{W}_{k}\boldsymbol{x}+\boldsymbol{b}_{k}
\end{equation}
where $\boldsymbol{W}_{k}$ and $\boldsymbol{b}_{k}$ are the learnable
parameters of the $k$-th mapping module $M_{k}$.

\subsection{Posterior Mapping Selection}

To ensure accurate optimization of mapping modules in training time, it is necessary to identify which mapping module is responsible for the generation of target response and only update the corresponding mapping module given the target response. 
Thus, we incorporate a posterior mapping selection module to explicitly select the corresponding mapping module by leveraging the information in target response. 
With the guidance of target response, we assume it is easier to find the corresponding mapping module for accurate optimization. 
Specifically, we introduce a categorical distribution $\pi$ to denote the selection of mapping module conditioned on the target response. 
And the selection probability of the $k$-th mapping module is based on its relevance to the target response, which is measured by the dot product between the representations of candidate response and target response as follows:
\begin{equation}
    \pi_{k}=\frac{exp(\boldsymbol{m}_{k}\cdot\boldsymbol{y})}{\sum_{i=1}^{K}exp(\boldsymbol{m}_{i}\cdot\boldsymbol{y})}
\end{equation}
Then for a target response, the corresponding mapping module can be sampled according to their relevance. 
Given that the $z$-th mapping module $M_{z}$ is selected,
only the corresponding candidate representation $\boldsymbol{m}_{z}$
is fed into the response decoder for further decoding optimization. 
Therefore, optimization of irrelevant mapping module is not conducted and more accurate optimization of latent mechanisms is ensured. In order to back-propagate through the discrete sampling in posterior mapping selection, we leverage the Gumbel-Softmax reparametrization \cite{JangGP17}.

\subsection{Decoder}

The response decoder employs an uni-directional GRU with the selected candidate representation $\boldsymbol{m}_z$ as its initial state and update its hidden state as follows:
\begin{equation}
    \boldsymbol{s}_{t}=\mathbf{GRU}(\boldsymbol{s}_{t-1},\boldsymbol{e}(y_{t-1}),\boldsymbol{c}_{t});\ \ \boldsymbol{s}_{0}=\boldsymbol{m}_z
\end{equation}
where $\boldsymbol{s}_{t}$ is the hidden state of decoder at time
$t$, $\boldsymbol{e}(y_{t-1})$ is the embedding of the last generated word $y_{t-1}$, and $\boldsymbol{c}_{t}$ is the attended context vector at time $t$
and defined as the weighted sum of hidden states of the post encoder: $\boldsymbol{c}_{t}=\sum_{i=1}^{T}\alpha_{t,i}\boldsymbol{h}_{i}$,
where $\alpha_{t,i}$ is the attention weight over $\boldsymbol{h}_{i}$
at time $t$:
\begin{align}
    \boldsymbol{\alpha}_{t} & =softmax(\boldsymbol{e}_{t})\\
    e_{t,i} & =\boldsymbol{v}^{T}tanh(\boldsymbol{W}_{h}\boldsymbol{h}_{i}+\boldsymbol{W}_{s}\boldsymbol{s}_{t})
\end{align}
where $\boldsymbol{v}$, $\boldsymbol{W}_{h}$ and $\boldsymbol{W}_{s}$ are learnable parameters. 
Then at time $t$, the generation probability conditioned on the input post $X$ and the selected mapping module $M_{z}$ is calculated as:
\begin{equation}
    p(y_{t}|y_{<t},X,M_{z})=softmax(\boldsymbol{s}_{t},\boldsymbol{c}_{t})
\end{equation}
where $y_{<t}$ denotes the previous generated words. 

The objective of the response generation is to minimize the negative log-likelihood of the target response $Y$ conditioned on the input post $X$ and the selected mapping module $M_{z}$ as follows:
\begin{equation}
    \mathcal{L}_{G}=-\ log\ p(Y|X,M_{z})
\end{equation}
where $p(Y|X,M_{z})=\prod_{t=1}^{T'}p(y_{t}|y_{<t},X,M_{z})$ is the conditional probability of target response.

\subsection{Auxiliary Objective}

Although the posterior mapping selection module is designed to select the corresponding mapping module by referring to the target response, we find that its raw implementation quickly converges to selecting the same mapping module and thus the proposed model falls back to the vanilla Seq2Seq.
We conjecture that in the early training, the response encoder is inefficient to capture the semantic information in target response. So the posterior mapping selection fails to provide an accurate selection of mapping module and the model falls into a local optima that focuses on single mapping module. 
To address this issue, we introduce an auxiliary objective from the response retrieval task to improve the semantic extraction of response encoder. 
The auxiliary objective namely \textit{matching loss} is to evaluate the relevance of post-response pair. Specifically, given an input post $X$ and a target response $Y$, their relevance probability is estimated by the dot product of their semantic representations:
\begin{equation}
    p(r=1|X,Y)=\sigma(\boldsymbol{x}\cdot\boldsymbol{y})
\end{equation}
where $r$ is the label denoting if the response $Y$ is relevant to the post $X$, and $\sigma$ is a sigmoid function. 
Following the previous work \cite{ShangFPFZY18}, we adopt negative sampling to train this auxiliary task so as to release the burden of human annotation. 
Particularly, for the input post $X$ and golden response $Y$, we randomly sample another response $Y^{-}$ in training set as a negative sample. 
Formally, the matching loss $\mathcal{L}_{M}$ is defined as the negative log-likelihood of relevance for the golden response and negative response:
\begin{equation}
    \mathcal{L}_{M}=-log\ p(r=1|X, Y)+log\ p(r=1|X,Y^{-})
\end{equation}
With this auxiliary matching loss, the response encoder is more efficient to capture the semantic information from the target response and provide a better relevance measurement for the accurate posterior mapping selection.

\subsection{Training and Generation}

Overall, the total loss function of our model is a combination of the generation loss $\mathcal{L}_{G}$ and the matching loss $\mathcal{L}_{M}$:
\begin{equation}
    \mathcal{L}=\mathcal{L}_{G}+\mathcal{L}_{M}
\end{equation}
All the parameters are simultaneously updated with back-propagation.

After optimization with posterior mapping selection, the model is able to capture distinct responding regularities and generate various candidate responses with different mapping modules. 
For response generation, we assume each mapping module is reasonable and just randomly pick mapping module for responding to avoid selection bias. 
More advanced response selection such as reranking is left as future work.

\section{Experiment}

\subsection{Datasets}

We evaluate the proposed model on two public conversation dataset:
\textbf{Weibo} \cite{ShangLL15} and \textbf{Reddit} \cite{ZhouYHZXZ18}
that maintain a large repository of post-response pairs from popular social websites. 
After basic data cleaning, we have above 2 million pairs in both datasets. 
The statistics of datasets are summarized in Table \ref{tab:statistics-of-datasets}.

\begin{table}
    \centering
    % \fontsize{7}{9}\selectfont
    \begin{tabular}{cccc}
        \hline 
        Dataset & \#train & \#valid & \#test\tabularnewline
        \hline 
        Weibo & 2,630,212 & 11,811 & 974\tabularnewline
        Reddit & 2,173,501 & 6,536 & 1,298\tabularnewline
        \hline 
    \end{tabular}
    \caption{Statistics of datasets.}
    \label{tab:statistics-of-datasets}
\end{table}

\subsection{Implementation Details}

%We use a vocabulary of 40,000 and 30,000 words in Weibo and Reddit dataset, respectively.
The vocabulary size is limited to 40,000 and 30,000 in Weibo and Reddit dataset, respectively.
%We take the most frequent 40,000 and 30,000 words as the vocabulary for Weibo and Reddit dataset, respectively. 
%And the remaining words are replaced by a special token ``UNK''.
% All hyper-parameters are set based on pilot experiments.
The hidden size in both encoder and decoder is set to 1024. 
Word embedding has size 300 and is shared for both encoder and decoder. 
We initialize the word embedding from pre-trained Weibo embedding \cite{LiZHLLD18} and GloVe embedding \cite{PenningtonSM14} for Weibo and Reddit dataset, respectively. 
The temperature of Gumbel-Softmax trick is set to 0.67.
All model are trained end-to-end by the Adam optimizer \cite{KingmaB15} on mini-batches of size 128, with learning rate 0.0002. 
We train our model in 10 epochs and keep the best model on the validation set for evaluation. \footnote{The code will be released at: \url{https://github.com/PaddlePaddle/models/tree/develop/PaddleNLP/Research/IJCAI2019-MMPMS}.}
%The code is available at: \url{https://github.com/LittleChencht/MMPMS}.

\subsection{Compared Methods}

We compare our model with several state-of-the-art generative conversation models in both single and multiple (i.e.\ 5) response generation:

\noindent \textbf{Seq2Seq }\cite{BahdanauCB15}: The standard Seq2Seq
architecture with attention mechanism.

\noindent \textbf{MMI-bidi} \cite{LiGBGD16}: The Seq2Seq model using
Maximum Mutual Information (MMI) between inputs and outputs as the objective function to reorder generated responses. 
We adopt the default setting: $\lambda=0.5$ and $\gamma=1$.

\noindent \textbf{CVAE} \cite{ZhaoZE17}: The Conditional Variational Auto-Encoder model with auxiliary bag-of-words loss.

\noindent \textbf{MARM} \cite{ZhouLCLCH17}: The Seq2Seq model augmented with mechanism embeddings to capture latent responding mechanisms.
We use 5 latent mechanisms and generate one response from each mechanism.

\noindent \textbf{MHAM} \cite{TaoGSWZY18}: The Seq2Seq model with multi-head attention mechanism. 
The number of heads is set to 5. 
Following the original setting, we combine all heads of attention to generate a response for the single response generation, and generate one response from each head of attention for the multiple response generation.
Although the constrained MHAM is reported better performance in the original paper, on our both datasets we see negligible improvement in single response generation and much worse performance in multiple response generation due to the lack of fluency. 
So we only adopt the unconstrained MHAM as baseline.

\noindent \textbf{MMPMS}: Our proposed model with \textit{\textbf{M}ulti-\textbf{M}apping} and \textit{\textbf{P}osterior \textbf{M}apping \textbf{S}election} (MMPMS). 
We set the number of mapping modules to 20.

\begin{table*}
    \centering
    \fontsize{8}{10}\selectfont
    \begin{tabular}{l|cccc|cccc}
        \hline 
        \multirow{2}{*}{Model} & \multicolumn{4}{c|}{Weibo} & \multicolumn{4}{c}{Reddit}\tabularnewline
        \cline{2-9} \cline{3-9} \cline{4-9} \cline{5-9} \cline{6-9} \cline{7-9} \cline{8-9} \cline{9-9} 
         & Acceptable & Good & BLEU-1/2 & Dist-1/2 & Acceptable & Good & BLEU-1/2 & Dist-1/2\tabularnewline
        \hline 
        Seq2Seq & 0.43 & 0.08 & \textbf{0.305/0.246} & 0.122/0.326 & 0.57 & 0.10 & 0.205/0.162 & 0.091/0.254\tabularnewline
        MMI-bidi & 0.46 & 0.09 & 0.271/0.218 & 0.153/0.372 & 0.54 & 0.25 & \textbf{0.345}/\textbf{0.279} & 0.107/0.325\tabularnewline
        CVAE & 0.29 & 0.15 & 0.252/0.203 & 0.184/0.542 & 0.42 & 0.25 & 0.287/0.233 & 0.107/0.428\tabularnewline
        MARM & 0.48 & 0.11 & 0.304/0.245 & 0.132/0.376 & 0.60 & 0.09 & 0.205/0.162 & 0.100/0.287\tabularnewline
        MHAM & 0.50 & 0.10 & 0.304/0.245 & 0.127/0.347 & 0.60 & 0.10 & 0.192/0.151 & 0.115/0.331\tabularnewline
        \hline 
        MMPMS & \textbf{0.56} & \textbf{0.24} & 0.275/0.225 & \textbf{0.189/0.553} & \textbf{0.65} & \textbf{0.36} & 0.207/0.165 & \textbf{0.135}/\textbf{0.433}\tabularnewline
        \hline 
    \end{tabular}{\small\par}
    \caption{Evaluation results of single response generation on Weibo and Reddit
    dataset.}
    \label{tab:results-single}
\end{table*}

\begin{table*}
    \centering
    \fontsize{8}{10}\selectfont
    \begin{tabular}{l|ccccc|ccccc}
        \hline 
        \multirow{2}{*}{Model} & \multicolumn{5}{c|}{Weibo} & \multicolumn{5}{c}{Reddit}\tabularnewline
        \cline{2-11} \cline{3-11} \cline{4-11} \cline{5-11} \cline{6-11} \cline{7-11} \cline{8-11} \cline{9-11} \cline{10-11} \cline{11-11} 
         & Acceptable & Good & Diversity & BLEU-1/2 & Dist-1/2 & Acceptable & Good & Diversity & BLEU-1/2 & Dist-1/2\tabularnewline
        \hline 
        Seq2Seq & 0.45 & 0.08 & 0.79 & 0.291/0.234 & 0.037/0.133 & 0.43 & 0.10 & 1.27 & 0.300/0.242 & 0.022/0.085\tabularnewline
        MMI-bidi & 0.47 & 0.09 & 0.99 & 0.272/0.219 & 0.047/0.166 & 0.52 & 0.24 & 1.45 & \textbf{0.339/0.274} & 0.028/0.127\tabularnewline
        CVAE & 0.22 & 0.13 & 1.06 & 0.275/0.224 & \textbf{0.105/0.404} & 0.42 & 0.27 & 2.06 & 0.279/0.226 & 0.047/0.257\tabularnewline
        MARM & 0.49 & 0.11 & 0.62 & \textbf{0.306/0.246} & 0.030/0.091 & 0.60 & 0.10 & 0.67 & 0.204/0.161 & 0.021/0.064\tabularnewline
        MHAM & 0.31 & 0.08 & 1.41 & 0.234/0.190 & 0.074/0.240 & 0.40 & 0.12 & 1.82 & 0.184/0.149 & 0.056/0.234\tabularnewline
        \hline 
        MMPMS & \textbf{0.60} & \textbf{0.25} & \textbf{2.61} & 0.270/0.219 & 0.100/0.389 & \textbf{0.68} & \textbf{0.38} & \textbf{3.16} & 0.195/0.159 & \textbf{0.060/0.265}\tabularnewline
        \hline 
    \end{tabular}
    \caption{Evaluation results of multiple response generation on Weibo and Reddit dataset.}
    \label{tab:results-multiple}
\end{table*}

\subsection{Evaluation Metrics}

For automatic evaluation, we report:
\textbf{BLEU} \cite{ChenC14}: A widely used metric for generative dialogue systems by measuring word overlap between the generated response and the ground truth.
 \textbf{Dist-1/2} \cite{LiGBGD16}: Ratio of distinct unigrams/bigrams
in the generated responses, which can measure the diversity of generated responses.

Since empirical experiments \cite{LiuLSNCP16} have shown weak correlation between automatic metrics and human annotation, we consider the careful human judgment as major measurement in the experiments.
In detail, three annotators are invited to evaluate the quality of the responses for 300 randomly sampled posts. 
Similar to \cite{ZhouLCLCH17}, for each response the annotators are asked to score its quality with the following criteria: 
\textbf{}(1)\textbf{ Bad}: The response is ungrammatical and irrelevant.
\textbf{}(2)\textbf{ Normal}: The response is basically grammatical and relevant to the input post but trivial and dull.
\textbf{}(3)\textbf{ Good}: The response is not only grammatical and semantically relevant to the input post, but also meaningful and informative. 
Responses on normal and good levels are treated as \textquotedblleft \textbf{Acceptable}\textquotedblright .
Additionally, to evaluate the diversity of multiple responses generation, for the 5 responses generated for a single post, the annotators also annotate the number of distinct meanings among the acceptable responses, namely \textbf{Diversity}. 
The average Fleiss\textquoteright{} kappa \cite{FleissJ73} value is 0.55 and 0.63 on Weibo and Reddit, respectively, indicating that the annotators reach moderate agreement.

\subsection{Evaluation Results}

The evaluation of single response generation are summarized in Table \ref{tab:results-single}. 
As shown, our model achieves the best performance in human evaluation and \textit{Dist} on both datasets, especially the visible enhancement in \textit{Good} ratio (0.24 vs.\ 0.15 on Weibo and 0.36 vs.\ 0.25 on Reddit) compared with the best baseline, indicating our model can generate more informative and diverse responses. 
Notably, Seq2Seq performs the best in BLEU but poor in human evaluation on Weibo, which further verifies the weak correlation of BLEU to human judgment.

Table \ref{tab:results-multiple} shows the evaluation results of multiple response generation. 
As can be seen, our model outperforms baseline methods by a large margin in human evaluation on both datasets. 
More importantly, the \textit{Diversity} measure of our model reaches 2.61 on Weibo and 3.16 on Reddit, much higher than other baseline methods, which demonstrates the superiority of our model to generate multiple diverse and high-quality responses. 
This can also be supported by the examples in Table \ref{tab:examples}\footnote{Due to space limitation, responses from Seq2Seq and MMI-bidi are omitted, considering their lack of one-to-many mechanism and low diversity in multiple response generation.}, where the multiple candidate responses returned by our model are much more relevant and diverse.

\begin{table*}[t]
    \centering
    \includegraphics[scale=0.75]{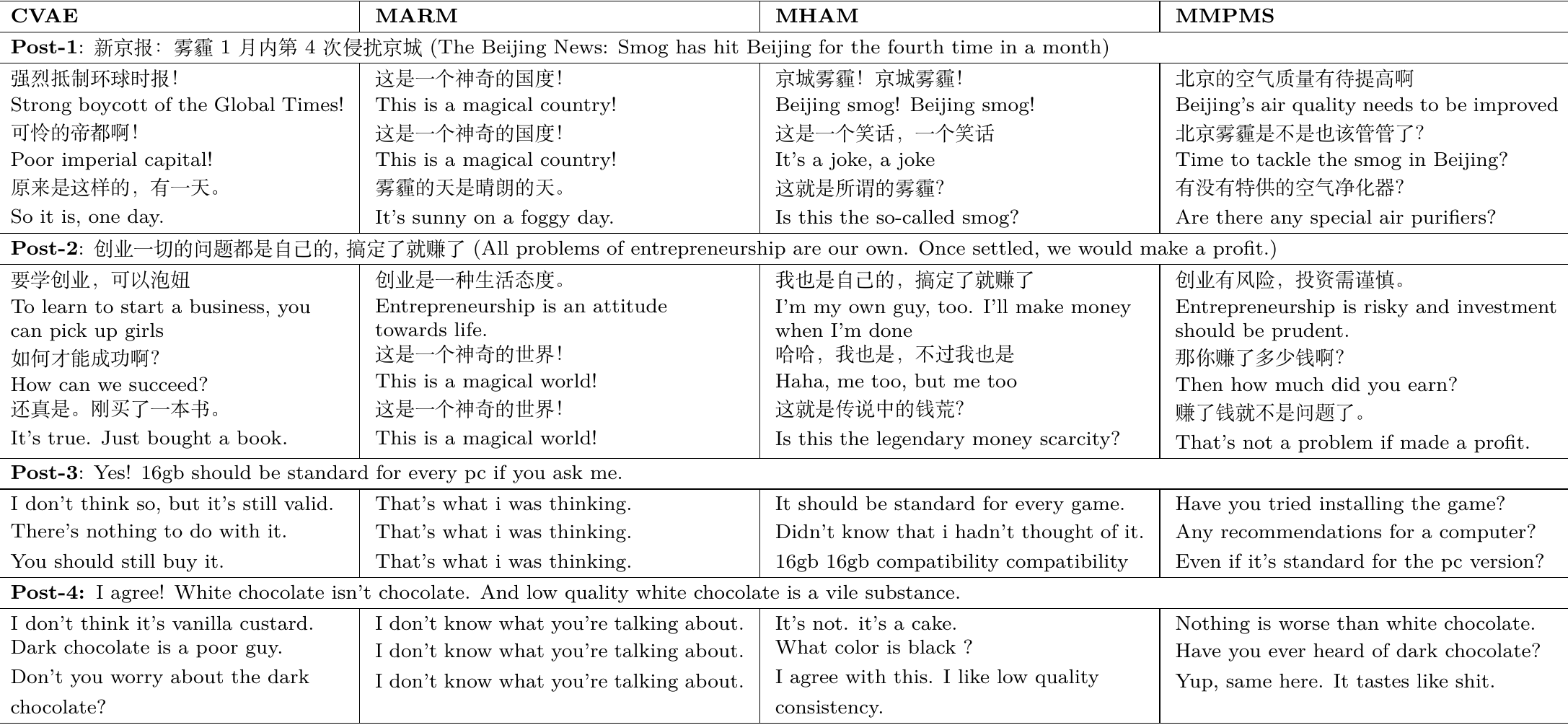}
    \caption{Examples of multiple response generation. The first two are from Weibo and the last two are from Reddit.}
    \label{tab:examples}
\end{table*}

However, CVAE fails to generate appropriate responses (i.e. low \textit{Acceptable} ratio) even though it achieves relatively high \textit{Dist} scores. 
It seems that its generation diversity comes more from the sampling randomness of the prior distribution rather than from the understanding of responding regularities. 
And we conjecture that the lack of multi-modal property in Gaussian distribution makes it hard to capture the one-to-many relationship among human conversation.

Instead, MARM performs the worst in response diversity with the lowest score of \textit{Diversity} and \textit{Dist} while it obtains a high \textit{Acceptable} ratio. 
As shown in Table \ref{tab:examples}, the responses from MARM are similar and trivial. And the word overlap among the 5 candidate responses is up to 94\% and 96\% on Weibo and Reddit, respectively, showing that each mechanism embedding converges to similar and general responding regularities.
We attribute this to the lack of accurate selection of latent mechanism for the target response during training. 
Since each mechanism embedding is roughly optimized with the same target response, they are prone to learn similar and general responding relationships. 
This result further validates the importance of our proposed posterior mapping selection.

It is also interesting to find the degradation of MHAM in multiple
response generation over single response generation. 
According to Table \ref{tab:examples}, the responses from different heads of attention are diverse but ungrammatical and irrelevant.
The reason may also lies in the absence of accurate selection of latent mechanism during training.
Since the response generation is optimized with a combination of all heads rather than the head corresponding to the target response, each head of attention is coupled together and fails to capture independent responding regularity. 
So the model ends up with inappropriate responses when only single head of attention is utilized in multiple response generation, but ends up with appropriate responses when combining all heads of attention in single response generation.
Another potential reason is that there is no enough distinct semantic information in the input post for the model to attend separately.

\subsection{Analysis on Mapping Modules}

\begin{figure}
    \centering
    \includegraphics[scale=0.33]{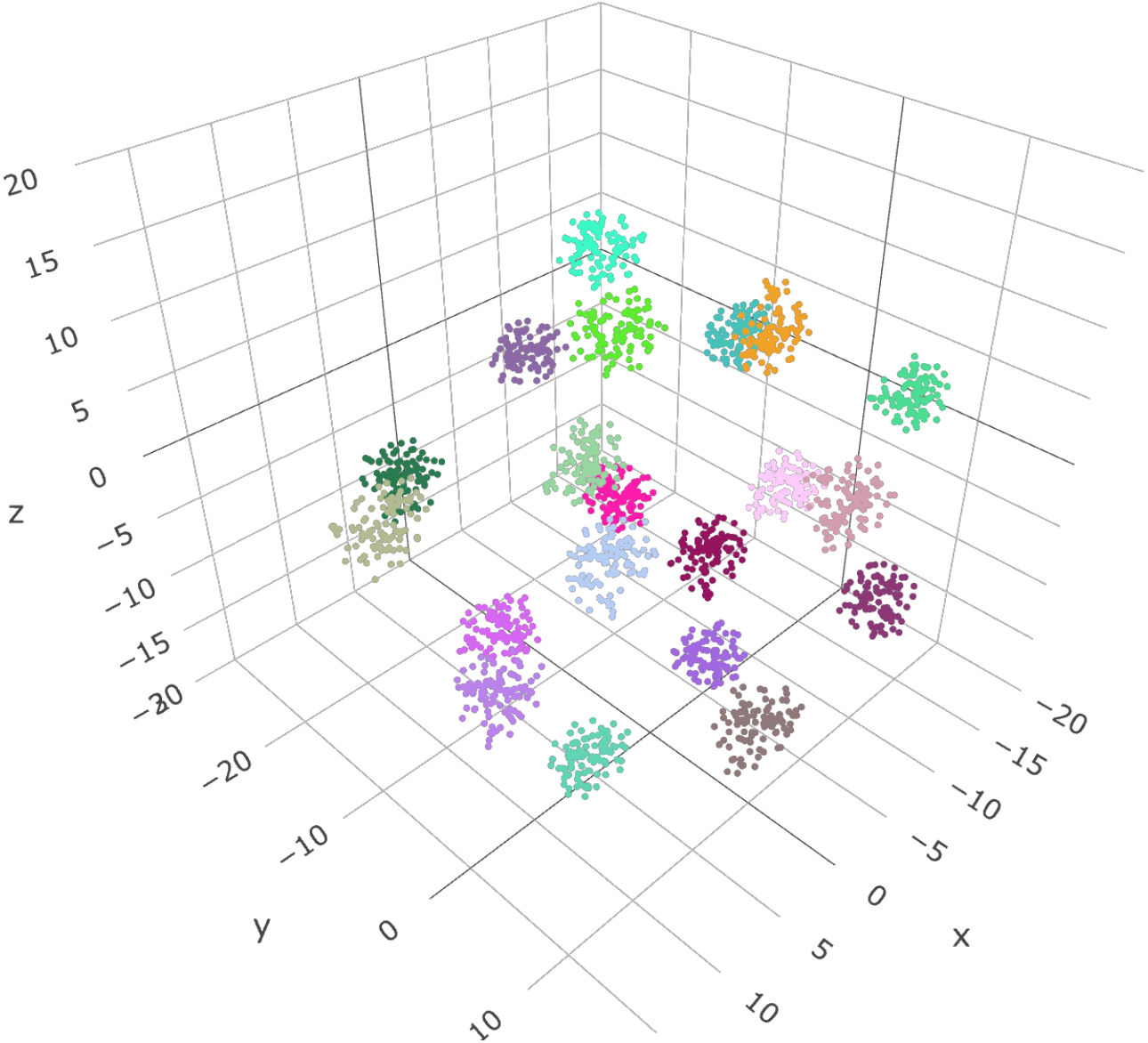}
    \caption{t-SNE visualization of candidate representations. The color represents the mapping module.}
    \label{fig:visualization}
\end{figure}

We also conduct analysis to explore the responding regularities that the mapping modules have captured. 
For the 200 posts sampled from the Weibo test set, we obtain the candidate representation $\boldsymbol{m}_{k}$ from different mapping modules and apply t-SNE \cite{MaatenH08} for visualization. 
As shown in Figure \ref{fig:visualization}, the candidate representations are highly clustered by their corresponding mapping modules, indicating the ability of various mapping modules to model various responding regularities by mapping the post representation to significantly different response representations.

For more intuitive understanding, we identify the keywords of different mapping modules from their responses for the Weibo test set. 
We assume that for a mapping module its keyword should appear frequently in its output responses but rarely occur in other mapping modules. 
Then, the importance of a word $w$ in the mapping module $M_{k}$ is measured by $p(w|M_{k})=N_{k}^{w}/\sum_{i=1}^{K}N_{i}^{w}$, where $N_{k}^{w}$ is the number of times that word $w$ occurs
in the responses from $M_{k}$. 
In addition, only the keywords occurring frequently are considered, namely $N_{k}^{w}>40$.
Table \ref{tab:keywords} illustrates the keywords of several representative mapping modules. 
As we can see, Map-1, whose keywords are mainly question words (e.g.\ \textit{where}, \textit{what} and the question mark), probably represents interrogative responding regularity. 
Map-2 tends to respond with mood words (e.g.\ \textit{wow}, \textit{lol}, \textit{haha}). 
The keywords of Map-3 are composed of intensity words (e.g.\ \textit{too}, \textit{so}) and modifier words (e.g. \textit{horrible}, \textit{cute}, \textit{pretty}), showing that it tends to respond with surprise and emphasis. 
And Map-4 is more likely to return responses in English. 
The keywords of Map-5 are mainly subjective words (e.g.\ \textit{I}, \textit{think}, \textit{believe}, \textit{want}), indicating that it tends to generate responses with respect to personal opinion or intention.
These results verify that each mapping module can capture diverse and meaningful responding regularities.

\begin{table}
    \centering
    \includegraphics[scale=0.70]{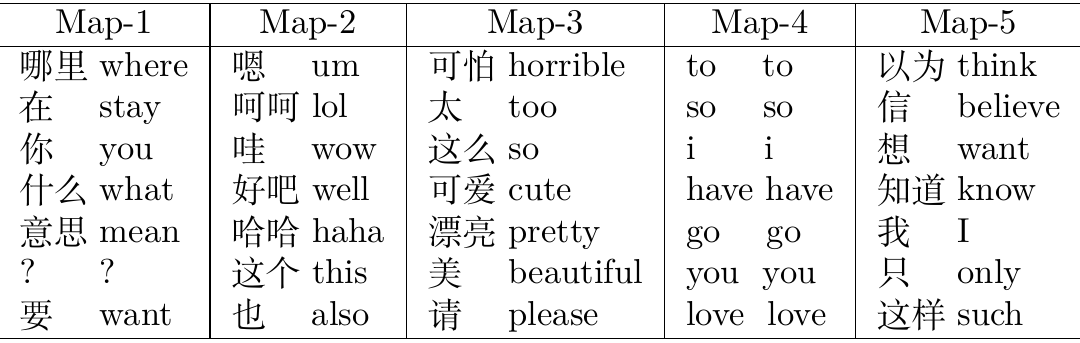}
    \caption{Keywords of different mapping modules.}
    \label{tab:keywords}
\end{table}

\section{Related Work}

The \textit{safe response} problem \cite{LiGBGD16} in Seq2Seq models \cite{SutskeverVL14,ShangLL15} remains an open challenge. 
In order to generate multiple diverse responses, many approaches resort to enhanced beam search \cite{LiGBGD16,LiMJ16}. 
But these methods are only applied to the decoding process and limited by the semantic similarity in the decoded responses. 
Another line of research turns to the different factors that determine the generation of diverse responses, such as sentence function \cite{HuangKGx18}, specificity \cite{ChengXGLZF18}, dialogue act \cite{XuWW18} and keywords \cite{GaoBLLS19}. 
However, such methods require annotations and can capture only one aspect of one-to-many relationship. 
Work from \cite{ZhaoZE17,SerbanSLCPCB17} combines Seq2Seq with Conditional Variational Auto-Encoder and employs a Gaussian distribution to capture discourse-level variations. 
But it is observed that these methods suffer from the \textit{posterior collapse} problem \cite{BowmanVVDJB16}. 
Moreover, the Gaussian distribution is not adaptive to the multi-modal nature of diverse responses.

The most relevant work to ours is those incorporating multiple latent mechanisms for one-to-many relationship. 
Zhou \textit{et al.}\ \shortcite{ZhouLCLCH17,ZhouLXLCH18} propose a mechanism-aware machine that introduces multiple latent embeddings as language responding mechanisms. 
Tao \textit{et al.}\ \shortcite{TaoGSWZY18} propose a multi-head attention mechanism to generate diverse responses by attending to various semantic parts of an input post. 
Gu \textit{et al.}\ \shortcite{gu2018dialogwae} incorporate a Gaussian mixture prior network and employ Gaussian component as the latent mechanism to capture the multi-modal nature of diverse responses.
However, without an accurate selection of the latent mechanism corresponding to the target response, these methods suffer from a rough optimization of latent mechanisms.
Given a target response, instead of optimizing the corresponding latent mechanism, Zhou \textit{et al.}\ \shortcite{ZhouLCLCH17} just roughly optimize each mechanism embedding while Tao \textit{et al.}\ \shortcite{TaoGSWZY18} optimize a rough combination of all heads of attention according to the input post.
Whereas, our model maintains a more accurate selection of the corresponding latent mechanism by referring to the posterior information in the target response.
Although posterior information is also utilized in \cite{gu2018dialogwae}, it is for the optimization of the prior Gaussian component which is roughly inferred by the input post, rather for the accurate selection of the Gaussian component.

\section{Conclusion}

In this paper, we augment the Seq2Seq model with a multi-mapping mechanism to learn the one-to-many relationship for multiple diverse response generation.
Particularly, our model incorporates a posterior mapping selection module to select the corresponding mapping module according to the target response for accurate optimization. 
An auxiliary matching loss is also proposed to facilitate the optimization of posterior mapping selection. 
Thus each mapping module is led to capture distinct responding regularities. 
Experiments and analysis support that our model works as expected and tends to generate responses of diversity and high quality.
%Our experiments and analysis support that the proposed model with multi-mapping and posterior mapping selection works as expected and tends to generate responses of diversity and high quality.

\section*{Acknowledgments}
We thank Rongzhong Lian, Siqi Bao and Huang He from Baidu Inc. for their constructive advice.
%We also thank the anonymous IJCAI reviewers for their helpful feedback.

\bibliographystyle{named}
\bibliography{reference}

\end{document}